\documentclass[conference]{IEEEtran}
\IEEEoverridecommandlockouts
\usepackage{graphicx}
\usepackage{amsmath}
\usepackage{xcolor}
\usepackage{booktabs}
\usepackage{placeins}
\usepackage{url}

\title{Anatomy of a Closed-Loop Collapse:\\
A Causal Case Study of a Compressed VLA Policy}

\author{\IEEEauthorblockN{Fengze Jia\thanks{Accepted as a poster at the IROS 2026 Workshop on Building Scalable Infrastructure for Robot Learning: From Data Scaling to Real-World Deployment (ScaleInfra), Pittsburgh, PA, USA. Supplementary records:\protect\\ \protect\url{https://github.com/Xanadum/closed-loop-collapse}}}
\IEEEauthorblockA{The Ohio State University\\
Columbus, OH, USA\\
jia.747@osu.edu}}

\begin{document}
\maketitle
\pagestyle{plain}\thispagestyle{plain}

\begin{abstract}
Compressed manipulation policies can pass offline evaluation while
failing in closed-loop execution; this offline--closed-loop
dissociation is established in prior work and is not our claim. We
contribute a causal anatomy of one naturally occurring case. An
8-layer distillation of Octo-Base retains 86\% of parameters, passes
every offline check we applied (0.996 and 1.000 teacher-ratios on the
model family's own validation metrics), and collapses in closed loop:
0/72 versus the teacher's 40/72 on a simulated WidowX pick-and-place
task. The collapse is structured, not diffuse: early task stages
degrade gradually (the student still moves the object at 90\% of
the teacher's rate and grasps at 55\%), while transport-to-target
fails categorically, at 0\% in every training variant. Paired
action-trace forensics isolate the signature: a negative, late-heavy
$z$ residual, roughly an order of magnitude larger than after
repair, and persistent across the base distillation and both
continuation branches. Four standard therapies fail under
matched controls: continued training and in-domain offline data leave success at zero, even though the latter measurably improves marginal action statistics; command-level compensation recovers nothing at any
offset, although the same perturbations demonstrably degrade healthy
policies; clamping the symptom in the command channel preserves
grasping, yet success stays at floor. A minimal-pair controlled intervention that substitutes half of the training stream with deployment-distribution teacher rollouts, with every other setting held fixed, restores parity with the teacher (18/36 vs.\ 17/36 held-out),
eliminates the negative, late-heavy signature, and recovers a
teacher-like perturbation-response profile. We claim existence, not
universality. The operational implication: offline gates, including a
family's own validation metrics, are insufficient acceptance tests
for compressed policies; a few dozen closed-loop trials sufficed to
find what they missed.
\end{abstract}

\section{Introduction}

We compressed a 12-layer generalist robot policy into an 8-layer
student and watched it pass every offline check we applied before
deployment. The compression is mild, retaining 86\% of parameters for a $1.08\times$ latency gain, and by the offline measures
of its own model family the student returns teacher-level ratios:
0.996 of the teacher's gripper-prediction accuracy, 1.000 of
its translation-direction error. In closed-loop execution of a
simulated WidowX pick-and-place task, the student still moves the
object in 18 of 24 episodes (teacher: 20) and grasps it in 11
(teacher: 20). It transports the object to the target in \textbf{zero}, with 0/24 on the instrumented slice, 0/72 on the full
protocol, and zero again in every retrained variant we produced. The
teacher scores 40/72 under an identical harness.

That gradient, from graded degradation through the early stages to a categorical cliff at the final one, is the object of this paper. The
coarse phenomenon behind it, strong offline metrics coexisting with
closed-loop failure, is not our discovery and we do not claim it: the
offline--closed-loop gap is established for manipulation policy
evaluation \cite{autoeval}, and its theoretical substrate,
compounding execution error under covariate shift, dates to
imitation-learning classics \cite{rossbagnell, dagger} with modern
treatments \cite{simchowitz}. What the literature does not contain,
to the best of a documented search (Sec.~\ref{sec:related}), is a
\emph{causal anatomy} of a naturally occurring instance: where such a
failure lives, which standard remedies fail against it and why, and
what suffices to cure it, all established at single-case resolution with
controlled interventions.

We contribute that anatomy:
\begin{enumerate}
  \item \textbf{A localized lesion.} Failure concentrates at one
  stage boundary (transport), and paired action traces isolate its signature, namely a negative, late-heavy $z$ residual, roughly
  $10\times$ its post-repair magnitude, persistent across the base
  distillation and both continuation branches
  (Sec.~\ref{sec:lesion}).
  \item \textbf{Controlled falsification of four standard
  therapies.} Continued training is inert. In-domain offline data is
  inert on success while measurably improving marginal action statistics, so whatever the deficit is, it is not those statistics.
  Command-level compensation recovers nothing at any offset, though
  the same perturbations demonstrably degrade healthy policies. The instrument works; the therapy does not; an additive-bias account is excluded. Clamping the symptom in the command channel preserves grasping and still fails, so the damage is not a single-channel
  defect (Sec.~\ref{sec:necessity}).
  \item \textbf{A sufficient therapy as a minimal pair, with
  completeness of recovery.} Substituting half the training stream
  with deployment-distribution teacher rollouts, while every other setting is held identical and verified from execution records, restores
  held-out success to teacher parity (18/36 vs.\ 17/36). Recovery is
  complete in a stronger sense: the negative, late-heavy trace
  signature is eliminated, and the repaired student exhibits a
  teacher-like perturbation-response profile, while the
  zero-rollout control stays at floor (Sec.~\ref{sec:sufficiency}).
\end{enumerate}

Our claim is existential, not universal. We do not assert that all
compressed VLAs fail, nor that all such failures share this anatomy.
We assert that this class of failure exists. It is mild-compression-induced, invisible to the standard offline battery, stage-localized, resistant to symptomatic and data-volume remedies, and curable by substituting successful teacher rollouts collected in the deployment environment into an otherwise unchanged training stream. We also assert that its anatomy can be established at single-case resolution with controlled interventions, the resolution existence claims require.

For evaluation infrastructure the consequence is operational. Six hundred held-out trajectories of offline evaluation returned teacher-level scores; twenty-four closed-loop rollouts separated the two policies completely, and one paired trace comparison isolated the failure's signature. Acceptance pipelines for compressed policies need closed-loop probes, and this case shows how little they can cost, since a few dozen trials sufficed here.

\section{Related Work}
\label{sec:related}

\textbf{Offline--closed-loop dissociation and evaluation.}
Prior manipulation-policy evaluations document that offline validation
metrics can provide unreliable estimates or rankings of closed-loop
performance \cite{autoeval}. Classical imitation-learning analyses
attribute such gaps to learner-induced state-distribution shift and
compounding execution error \cite{rossbagnell, dagger}, with a recent
continuous-action treatment in \cite{simchowitz}. The nearest prior in
vocabulary, Factored Scaling Curves \cite{fsc}, shows that a
purpose-built offline proxy, namely policy-embedding similarity, can effectively rank data-collection choices, including for generalization
to unseen environment variations; ranking data-collection options and
accepting a compressed policy for deployment are \emph{different
decisions}, and our case concerns the latter. We do not revisit the
existence of this gap or its theoretical account; we contribute the
causal anatomy of one naturally occurring, compression-induced
instance.

\textbf{VLA compression and distillation.}
A growing methods literature compresses VLA policies via architectural
miniaturization \cite{tinyvla}, early-exit decoding \cite{ceedvla},
action-expert and self-derived distillation \cite{vitavla,actdistill},
pruning-based recovery pipelines \cite{rlrc}, and VLM-assisted
distillation \cite{vlaad}. These works deliver methods, that is, smaller policies that work, whereas ours delivers a causal anatomy of why a compressed
policy that should work does not. Two of them parallel our anatomy
from the healthy side of the boundary. On the necessity
side, \cite{rlrc} reports the same shape at a different depth:
structured pruning collapses closed-loop success to zero, supervised
fine-tuning on offline demonstrations recovers much but not all of it,
and only on-policy interaction in the deployment environment completes
recovery to parity. On the sufficiency side, the pipeline of
\cite{vlaad} trains its student on teacher rollouts executed in the deployment environment, which closely matches the data regime our controlled repair identifies as sufficient, and flags success-only
filtering as an untested limitation, the same confound we disclose in
Sec.~VII.

\textbf{Failure analysis and error dynamics.}
SilentDrift \cite{silentdrift} demonstrates a closely related error
dynamic by attack: smooth, phase-targeted micro-perturbations
implanted through training-data poisoning compound through chunked
open-loop execution into task failure. That dynamic is engineered there, whereas our compression case arises naturally. Diagnostic
work dissects VLA policies at other levels: representation- and
pathway-level tracing attributes control to modality pathways in
intact models \cite{vlatrace}; mechanistic studies localize function
inside VLA internals via activation injection and sparse autoencoders
\cite{notallfeatures}; counterfactual failure synthesis generates
failure--correction pairs from successful demonstrations to train
recovery \cite{cfs}; and gripper-state ambiguity analysis root-causes
one specific failure mechanism and repairs it \cite{gripperstate}.
None jointly combines stage-level localization, action-level marker
identification, and interventional testing of competing explanations
for a naturally occurring, compression-induced closed-loop failure, the anatomy this paper contributes.

\textbf{Novelty statement.}
Under a documented retrieval protocol with three channels (faceted
keyword search, citation-graph traversal from seven verified seed
papers, author tracking) that yielded 1{,}001 deduplicated candidates with tiered full-text review of top-risk items, we found no prior
work satisfying the criteria above jointly. 

\section{The Case and Evaluation Protocol}
\label{sec:case}

\subsection{The compressed policy}
The teacher is Octo-Base-1.5 (12 transformer layers) \cite{octo}.
The student inherits 8 layers by expert surgery, which retains layers $\{0,1,3,4,6,8,10,11\}$ (both boundary layers plus a uniform interior sample), and trains for 300k steps (batch 128, peak LR
$3{\times}10^{-5}$ reached after 2{,}000 warmup steps,
cosine-decayed to $10^{-6}$) on the Open
X-Embodiment mixture \cite{oxe}, with a convex combination of
hidden-state distillation (weight $\alpha$; cosine distance over
observation, task, and readout token groups) and the diffusion
action head's native loss (weight $1{-}\alpha$), at $\alpha{=}0.5$.
Training telemetry is stable through 300k steps:
over the final 10k steps the distillation loss holds at
$\approx 0.010$ under the $1-\cos$ definition, action loss averages
$0.98$ (range $0.81$--$1.23$ across per-100-step records), and
gradient norm averages $4.3$ ($3.6$--$5.3$). Throughout,
``compression'' abbreviates this complete procedure, comprising layer removal, layer selection, the distillation
objective, and the ensuing optimization; its components are never
separated in this study.

The compression is deliberately mild, and we state its economics
against our own interest: total parameters 202.5M $\rightarrow$
174.1M (86\% retained; the transformer stack itself compresses to
68\%, diluted by shared tokenizers), mean inference latency 135.4
$\rightarrow$ 124.9\,ms ($1.08\times$). We do not advocate this
configuration as an efficiency method; it is the object of study.
That so mild a compression, under so unremarkable an offline
profile, produces a categorical closed-loop failure is precisely
what makes the case informative.

\subsection{Offline evaluation: the acceptance gates}
Offline evaluation uses the model family's own validation utilities
over held-out trajectories from the three in-distribution datasets of
the training mixture (Bridge, Fractal, Kuka; 200 each):
gripper-prediction accuracy and end-effector translation-direction
error, with a stricter auxiliary translation metric excluded as
uninformative ($<1\%$ for both models). The student passes at
teacher-ratios of 0.996 (gripping) and 1.000 ($xyz$ angle); ratios
are oriented so that 1.0 denotes teacher parity for accuracy and
error metrics alike. By the gates a practitioner would apply before
deployment, this distillation succeeded.

A first-pass retention sweep extends the same metrics to three
exploratory datasets outside the training mixture (fridge-opening,
tabletop rotation, pouring; 200 each). Across all six probes the
student tracks the teacher with small, sign-mixed differences, ahead on four of the six direction-error readings, and on the battery's
best-scoring probe, fridge-opening, it matches the teacher outright:
$13.3^\circ$ direction error at 0.914 strict accuracy, against the
teacher's $14.2^\circ$ at 0.918. Nothing in this battery
distinguishes the student from a successful distillation. We use
these readings as what they operationally are, namely the standard battery's acceptance verdict, and make no offline fidelity claim
beyond gate passage; per-dataset records are provided in full in
the supplementary material.\footnote{\url{https://github.com/Xanadum/closed-loop-collapse}}

\subsection{Closed-loop protocol}
Closed-loop evaluation runs in SimplerEnv / ManiSkill2 (SAPIEN)
\cite{simpler, maniskill2} on \texttt{PutEggplantInBasketScene-v0}
with a WidowX arm. Gripper actions binarize at threshold 0.75, a calibration for the Octo-1.5 model family. It was verified by a sanity run of the older Octo base and is applied identically to every policy here.
The diagnostic protocol evaluates 72 configurations (seeds
$\{0,1,2\}\times$ episode IDs 0--23); a single-seed slice carries
dense instrumentation. Staged outcomes are logged by the
environment's own evaluator: object moved, grasped, sustained hold
($\geq$5 consecutive steps), transport to target.\footnote{Exact
predicates extracted from the environment source are provided in the
metric-definitions document of the supplementary
material.} Repeatability: the teacher scores 15/24 and 14/24 on
identical reruns; we attribute $\pm1$ to GPU floating-point
nondeterminism over 120-step rollouts. All $z$ quantities in this paper, including paired residuals, injected offsets, and clamp floors, are per-step world-frame translation commands in meters, measured after
action unnormalization and identical in scale; $+z$ denotes upward
motion.

For the repair phase, episode IDs partition into a
selection set ($\{0$--$2, 9$--$14, 21$--$23\}$) and a held-out
set ($\{3$--$8, 15$--$20\}$; 12 IDs $\times$ seeds $\{0,1,2\}$ = 36
configurations). Checkpoint selection used only selection-set
configurations, and was locked before any held-out evaluation; the
held-out set was then evaluated exactly once per checkpoint for three
bracketing checkpoints. Checkpoint identity is enforced by per-step
staging directories (guarding a loader default that silently resolves
to the latest step), and action-normalization statistics are verified
bit-identical across all compared checkpoints ($\max|\Delta|=0$).

\subsection{Evidence grades}
Three measurement grades recur below, with different load ratings.
\emph{Counted outcomes} (36- or 24-trial success and stage counts)
are robust and carry every causal claim. \emph{Paired trace
statistics} (per-episode student$-$teacher differences, episode
composition held fixed) control initialization-level between-episode variation, improving sensitivity at small $n$, and carry
panel-scoped, action-level diagnostic claims. \emph{Unpaired
micro-displacement differences} at the 0.002--0.006 scale are
dominated by natural per-episode spread ($\pm0.003$) at attainable
$n$ and carry \emph{no} claims in this paper; where relevant we say
so explicitly (Sec.~\ref{sec:necessity}). This discipline is itself
a finding we commend to evaluation-infrastructure practice.

\section{The Lesion: A Graded Collapse with a Categorical Edge}
\label{sec:lesion}

\begin{table}[t]
\caption{Staged closed-loop outcomes, instrumented slice (seed 0,
episodes 0--23). Stages are the environment evaluator's own
predicates.}
\label{tab:closedloop}
\centering
\begin{tabular}{lccccc}
\toprule
Policy & Succ. & Moved & Grasp & Sust. & Target \\
\midrule
Teacher (12L)            & 15/24 & 20 & 20 & 19 & 15 \\
Student V1 (8L, 300k)    & 0/24  & 18 & 11 & 11 & 0  \\
\;+5k OXE (Step 0)       & 0/24  & 19 & 16 & 13 & 0  \\
\;+5k Bridge (Step 1)    & 0/24  & 20 & 12 & 10 & 0  \\
\bottomrule
\end{tabular}
\end{table}

Table~\ref{tab:closedloop} contains the paper's founding
observation, and it is a shape, not a number. Across the first three
stages the student is a degraded but recognizable copy of its teacher, moving the object at 90\% of the teacher's rate, grasping
at 55\%, holding at 58\%. At the final stage the copy ends: transport
to target is 0/24, and it is 0 in \emph{every} variant, with 0/72 on the full protocol and zero again after each retraining of
Sec.~\ref{sec:necessity}. The early stages degrade gradedly; the
last fails categorically. A policy that has lost capability
uniformly does not look like this; a policy whose conditional
mapping is intact on some state regions and deficient on others
does.

The dissociation also survives independent re-measurement: across
full re-evaluations the teacher scores 40/72 (diagnostic protocol)
and 36/72 (split protocol), while the student never exceeded 2/36
in any evaluation run of this project (0/72 diagnostic; 0/36
held-out; 1/36 on a selection-set anchor; 1--2/36 across
perturbation sweeps). We disclose every student non-zero we
observed; none approaches the teacher's band.

Paired action-trace analysis identifies a persistent residual
signature on a fixed diagnostic panel. The panel comprises four
distinct hard initializations (IDs 2, 11, 14, 23), retained from a
six-episode list (IDs 2, 11, 14, 15, 20, 23) fixed in the
drift-instrumentation scripts before any paired-trace analysis
existed: the four on which the teacher
also fails, matching failure status across policies (all four lie
in the selection set). The panel was held fixed for every
subsequent model comparison; V1 reaches grasp on 2 of 4.
Comparisons are therefore matched by episode initialization,
although the policies may occupy different states after their
closed-loop trajectories diverge. Per-(episode, time-third)
differencing against the teacher controls for initialization-level
between-episode variation and improves sensitivity at $n=4$; it
does not make four episodes representative. We therefore scope
all trace claims to this panel, while every causal claim rests on
counted outcomes elsewhere.

On this panel, the mean student--teacher $z$ residual is negative
in each time-third and largest in the final third of the 120-step
rollout: mean $d_z=-0.0067$ over the full rollout and
$-0.0103$ in the final third for V1. These intervals are fixed
thirds by step count, not grasp-event-aligned phases. Because V1
grasps on only 2 of 4 panel episodes, an event-aligned post-grasp
analysis would rest on at most two episodes. We therefore describe
the marker temporally as late-heavy and let the counted stage
profile in Table~\ref{tab:closedloop} establish the link to the
transport transition.

The selected marker is a negative, late-heavy $z$ residual,
roughly an order of magnitude larger than the repaired model's
residual (Sec.~\ref{sec:sufficiency}). It persists in both
continuation branches on the same panel
($-0.0067/-0.0048/-0.0047$ over the full rollout;
$-0.0103/-0.0073/-0.0081$ in the final third), despite
continued training under two different data regimes. It is
therefore a persistent pattern across these branches rather than
an idiosyncrasy of one checkpoint. Its role is diagnostic, not
causal: on this fixed panel it separates the three failing training
variants from the repaired checkpoint. Section~\ref{sec:necessity}
then tests two simple causal accounts directly, namely a fixed additive
$z$ bias and a single-channel $z$ defect.

Why $z$? Not by magnitude, because the gripper-open marginal difference
is larger. Two counted observations narrow the field first, each
stated at its precise strength. Conditional grasp retention is
largely preserved once a grasp is registered: all 11 of V1's
grasped episodes reach the sustained-hold criterion, as do 13 of
16 and 10 of 12 for the two continuation branches. Because the
flag is latched, this establishes retention for at least 5
consecutive steps but says nothing about state-conditional misuse
later in the episode. Yet no episode reaches the target in any
variant. The categorical break therefore lies between sustained
holding and target arrival, rather than in the mere ability to
establish and briefly retain a grasp.

Step~1 (Sec.~\ref{sec:necessity}) also changes the observed
gripper and rotation marginals substantially while success remains
at zero. This shows that reducing those marginal discrepancies is
insufficient for recovery; it does not exclude state-conditional
defects in either channel, consistent with the later evidence that
the damage is not confined to one action dimension.

Among the measured action-level candidates, $z$ is prioritized by
conjunction: negative early-, late-, and full-rollout phase means,
late-heavy timing, persistence across the base distillation and
both continuation branches, and mechanistic relevance, since a negative teacher-relative $z$ residual is compatible with insufficient
upward lift during transport. These are criteria for selecting an
intervention target, not proof of causality. The interventions of
Sec.~\ref{sec:necessity} subsequently demote even $z$ itself: it is
the most visible projection of the damage, not the damage. The
closure evidence is convergent rather than independent. Under the
repair of Sec.~\ref{sec:sufficiency}, the early-, late-, and
full-rollout residuals drop to near zero on this same panel, while
held-out success returns under the counted protocol. The two
measurements move together across model conditions.

\begin{table}[t]
\caption{Paired $z$ residual on the fixed four-episode diagnostic
panel (IDs 2, 11, 14, 23). Entries are mean student$-$teacher
$z$ differences, computed per episode and then averaged
($n{=}4$ episodes per cell); intervals are fixed thirds of the
120-step rollout. Negative values indicate the student commands
less upward displacement than the teacher on the same
initialization. The base distillation and both continuation
branches share the residual's sign in the early-third, final-third,
and full-rollout measures, along with its late-heavy temporal
shape; under repair these residuals drop to near zero and the
negative, late-heavy pattern is eliminated.}
\label{tab:signature}
\centering
\begin{tabular}{lrrrr}
\toprule
Policy & Early & Mid & Late & Full \\
\midrule
Student V1 (8L, 300k)   & $-0.0069$ & $-0.0028$ & $-0.0103$ & $-0.0067$ \\
\;+5k OXE (Step 0)      & $-0.0064$ & $-0.0007$ & $-0.0073$ & $-0.0048$ \\
\;+5k Bridge (Step 1)   & $-0.0061$ & $+0.0001$ & $-0.0081$ & $-0.0047$ \\
\midrule
Repaired (step 8k)      & $-0.0005$ & $+0.0037$ & $-0.0007$ & $+0.0009$ \\
\bottomrule
\end{tabular}
\end{table}

\section{Falsified Therapies: The Necessity Arm}
\label{sec:necessity}

A causal claim earns its keep against alternatives. We admit a
hypothesis when it has mechanistic plausibility, entanglement with
the target variable in our data, and intervenability;
Table~\ref{tab:hypotheses} lists every admitted hypothesis with its
decisive test. This section walks four therapies of increasing
intervention depth, each targeting one hypothesis, each failing
informatively. All verdicts below rest on counted outcomes.

\begin{table}[t]
\caption{Hypothesis ledger. Admission: plausibility $\times$
entanglement $\times$ intervenability.}
\label{tab:hypotheses}
\centering
\footnotesize
\begin{tabular}{p{2.35cm}p{2.5cm}p{2.1cm}}
\toprule
Hypothesis & Decisive test & Verdict \\
\midrule
H1 Same-data continuation is sufficient &
Step 0: continue, same data &
Not supported at tested budgets (0/24; 0/36) \\
H2 Tested offline task-domain augmentation is sufficient &
Step 1: add Bridge data &
Insufficient (marginals move; success stays 0/24) \\
H3 Fixed additive $z$ bias in the command channel &
$z$-offset sweep &
Excluded over tested offsets (no recovery; healthy policies degrade) \\
H4 Suppressing the negative $z$ command is sufficient &
Phase-gated command clamp &
Falsified (grasp retained; success at floor) \\
H5 State-conditional policy deficiency in the deployment regime &
Rollout-data minimal pair (Sec.~VI) &
Supported (bundle sufficient; active component unresolved) \\
H6 Failure solely from simulator physics &
Teacher control; in-sim repair &
Constrained (Sec.~VII) \\
H7 8L capacity ceiling on this task &
Same 8L reaches post-repair parity &
Falsified for this task \\
H8 Audited harness defect explains the gap &
Positive control (teacher 40/72); repeatability; audits &
Excluded within audited components \\
\bottomrule
\end{tabular}
\end{table}

\subsection{Continued training (H1)}
Step 0 resumes the student from 300k for 5{,}000 steps on the
unchanged OXE mixture (peak LR $3{\times}10^{-6}$): the control for
``it just needed more training.'' Success 0/24 (moved 19, grasped
16, sustained 13, target 0); the paired signature persists
($d_z=-0.0048$ all-phase). A second, longer control appears in
Sec.~\ref{sec:sufficiency}: ten thousand further steps of pure OXE
under the repair-phase settings also end at 0/36.

\subsection{In-domain offline data (H2)}
Step 1 is the folk remedy: if the task domain is under-covered, add
its data. A dual-stream schedule interleaves the OXE mixture with
the task-domain Bridge dataset (repair batch 64 every third step,
weight 0.25) for 5{,}000 steps from the same fork. The therapy \emph{visibly moves marginal action statistics}. The gripper close-fraction gap shrinks from $-0.161$ to $-0.065$, raw-open from $0.097$ to $0.046$, and rotation from $0.0053$ to $0.0033$ (paired
differences on the six-episode instrumentation list of
Sec.~\ref{sec:lesion}, collected before the repair-phase split was defined). Success stays at 0/24 with transport still zero.
Improving the observed action marginals is therefore insufficient
for recovery, and frames from the task-domain dataset are not, by
themselves, what is missing. Whether the broader
successful-teacher-rollout bundle supplies supervision that this
task-domain augmentation does not is the question
Sec.~\ref{sec:sufficiency} tests.

\subsection{Command-level compensation (H3)}
The signature's small, constant-sign, concentrated shape is what an additive-bias account predicts, and that
account would deflate this paper to a calibration footnote: add
$+\delta z$ at deployment and be done. It is the most deflationary
competitor and must be tested first. We sweep fixed additive
$z$-corrections over $\{0.002, 0.004, 0.006, 0.008, 0.012, 0.020\}$,
applied to every commanded action throughout the rollout, 36
counted trials per offset.

No offset recovers the student: success reads $1, 0, 1, 0, 0, 0, 0$
across the grid (baseline included), within the student's observed
floor range (0--2/36). The null is not an artifact of an inert
instrument: the same perturbations, applied to healthy policies
(Sec.~\ref{sec:sufficiency}, Table~\ref{tab:dose}), degrade them
monotonically; at offsets $\geq 0.008$ the persistent upward bias
prevents grasp formation for every policy. The instrument
demonstrably alters behavior; it cannot rescue this policy at any
dose. An additive bias, a correctable constant riding on an otherwise-correct policy, is excluded.

One honest limit. We attempted to quantify \emph{how} the policy's
own commands respond to injected offsets at the trace level.
At the injected scale (0.002--0.006) the per-episode natural spread
of post-grasp $z$ ($\pm0.003$) dominates the signal, and grasp
stochasticity varies the episode composition across conditions at
attainable $n$ (2--6). These measurements are included in the
supplementary material but carry no mechanism claims here. This is an instance of the evidence-grade discipline of
Sec.~\ref{sec:case}.

\subsection{Command-level clamping (H4)}
The offset sweep excludes only the simplest command-space account:
a fixed additive $z$ bias riding on an otherwise correct policy. It
does not determine whether the negative $z$ command itself is an
operative part of the failure, because the policy's subsequent
closed-loop commands can counteract a constant correction. We
therefore replace the correction with a constraint that cannot be
counteracted in the command space: a phase-gated hook clamps the
commanded $z$-delta to a floor, engaging once the policy first
issues a gripper-close command, a command-level phase marker that
precedes, and does not require, the evaluator's physical grasp
predicate. The gate works: grasping is preserved (16--17/36) and
the commanded net $z$-delta after gate engagement turns
non-negative by construction ($-0.0066 \to +0.0012/+0.0031$).
Success does not return: 2/36 at the unclamped base, 0/36 and 2/36
under the two clamp settings, the student's floor. Trajectories
shift laterally (mean commanded $x$ displacement after gate
engagement roughly doubles; small-$n$ trace observation, reported
qualitatively). With the headline symptom forbidden in the command
channel, the policy still cannot transport. The damage is not a
single-channel $z$ defect; the drift is the most visible projection
of a deficient mapping, not the deficiency itself.

\subsection{The intervention-depth gradient}
\begin{figure}[t]
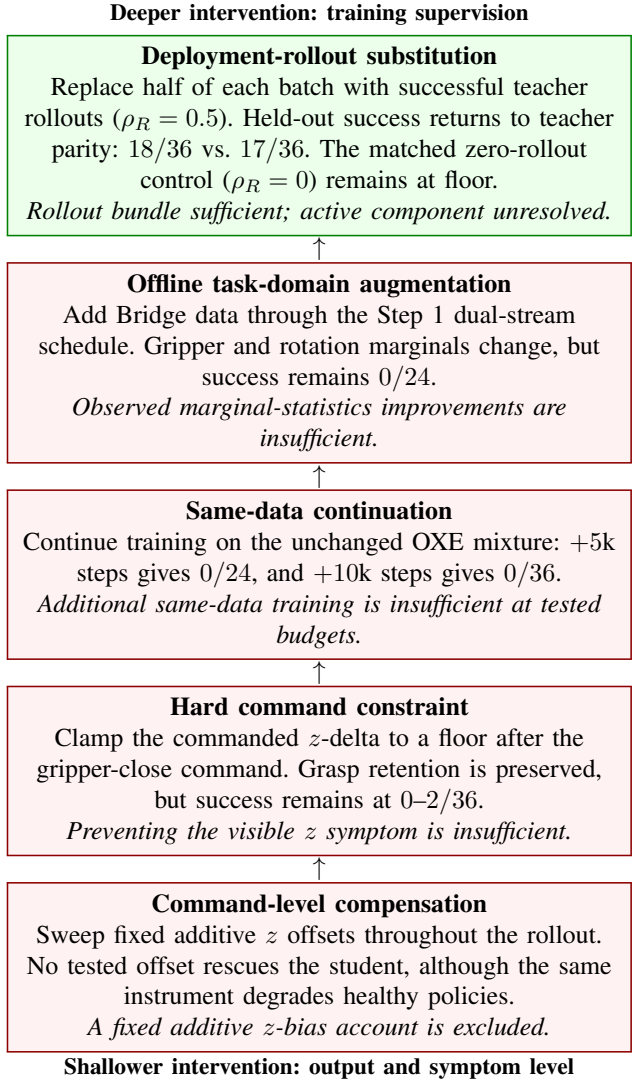

\centering
\setlength{\fboxsep}{4pt}
\setlength{\fboxrule}{0.6pt}
\resizebox{0.98\columnwidth}{!}{%
\begin{tabular}{c}
\textbf{\small Deeper intervention: training supervision}
\\[2pt]
\fcolorbox{green!50!black}{green!8}{%
\parbox{8.2cm}{\centering
\textbf{Deployment-rollout substitution}\\
Replace half of each batch with successful teacher rollouts
($\rho_R=0.5$). Held-out success returns to teacher parity:
$18/36$ vs.\ $17/36$. The matched zero-rollout control
($\rho_R=0$) remains at floor.\\
\textit{Rollout bundle sufficient; active component unresolved.}
}}
\\[-1pt]
$\uparrow$
\\[-1pt]
\fcolorbox{red!55!black}{red!5}{%
\parbox{8.2cm}{\centering
\textbf{Offline task-domain augmentation}\\
Add Bridge data through the Step~1 dual-stream schedule.
Gripper and rotation marginals change, but success remains
$0/24$.\\
\textit{Observed marginal-statistics improvements are insufficient.}
}}
\\[-1pt]
$\uparrow$
\\[-1pt]
\fcolorbox{red!55!black}{red!5}{%
\parbox{8.2cm}{\centering
\textbf{Same-data continuation}\\
Continue training on the unchanged OXE mixture:
$+5$k steps gives $0/24$, and $+10$k steps gives $0/36$.\\
\textit{Additional same-data training is insufficient at tested budgets.}
}}
\\[-1pt]
$\uparrow$
\\[-1pt]
\fcolorbox{red!55!black}{red!5}{%
\parbox{8.2cm}{\centering
\textbf{Hard command constraint}\\
Clamp the commanded $z$-delta to a floor after the
gripper-close command.
Grasp retention is preserved, but success remains at
$0$--$2/36$.\\
\textit{Preventing the visible $z$ symptom is insufficient.}
}}
\\[-1pt]
$\uparrow$
\\[-1pt]
\fcolorbox{red!55!black}{red!5}{%
\parbox{8.2cm}{\centering
\textbf{Command-level compensation}\\
Sweep fixed additive $z$ offsets throughout the rollout.
No tested offset rescues the student, although the same
instrument degrades healthy policies.\\
\textit{A fixed additive $z$-bias account is excluded.}
}}
\\[2pt]
\textbf{\small Shallower intervention: output and symptom level}
\end{tabular}%
}
\caption{\textbf{The therapy ladder.}
Interventions are ordered from shallow output-level corrections at
the bottom to upstream changes in training supervision at the top.
The four lower rungs fail under counted closed-loop trials, ruling
out progressively richer simple explanations. Only the top rung
restores teacher parity ($18/36$ vs.\ $17/36$ held-out); its
matched zero-rollout control stays at floor, attributing the
recovery to the substituted data rather than the additional
optimization. This establishes the rollout-data bundle as
sufficient for recovery, not any individual component within that
bundle.}
\label{fig:ladder}
\end{figure}

Read as a ladder (Fig.~\ref{fig:ladder}): command-level correction
fails at every dose; the hard command constraint is circumvented;
retraining on the unchanged mixture is inert at 5k and at 10k
steps; retraining with offline task-domain data is inert on
success while active on marginals. All four rungs fail. The arrow
points upstream, from output-level symptom correction into the
training-time supervision that formed the conditional mapping;
Sec.~\ref{sec:sufficiency} intervenes there with a matched
substitution of the successful-teacher-rollout bundle.

\section{Sufficient Therapy under Minimal-Pair Control}
\label{sec:sufficiency}

\subsection{Deployment-distribution data}
We collect 324 successful teacher rollouts in the deployment
environment via environment-variable-gated recording hooks, under
configuration-level isolation: recording uses the selection-set
episode IDs ($\{0$--$2, 9$--$14, 21$--$23\}$) under recording
seeds $\{100$--$179\}$, disjoint from the evaluation seeds
$\{0,1,2\}$. The seed governs only the
policy's action-sampling stream, while the initial scene is a
deterministic function of the episode ID alone; hence no recorded
(seed, ID) pair coincides with any of the 72 diagnostic
configurations, and the held-out episode IDs
($\{3$--$8, 15$--$20\}$) are never recorded under any seed. Episodes
are packaged as an RLDS dataset (38{,}880 frames; the count is
independently confirmed by the trainer's own configuration record)
and verified end to end: whole-stack image fingerprinting confirms
324/324 unique episodes with zero round-trip error;
action-normalization statistics are bit-identical to the student's
($\max|\Delta|=0$).

\subsection{The minimal pair, from execution records}
Two runs of the \emph{same} training script resume from the V1 300k
checkpoint. Their execution records (experiment-tracking config and
command-line metadata, included in the supplementary
material) agree on every setting, namely seed (42), learning rate (constant $10^{-6}$, warmup 200, no decay),
optimizer state reset, loss (the original distillation objective at
$\alpha{=}0.5$; no task-specific or $z$-specific terms added), total
batch (128), steps (10{,}000), checkpoint cadence (every 500), fork point, host, and GPU. They differ in exactly one argument:
$\rho_R = 0.5$ versus $\rho_R = 0$, i.e., whether half of each batch
is drawn from the rollout dataset (64+64) or none is (128+0).

Two properties of this design carry weight. First, it is
\emph{substitution, not addition}: the repair run receives no extra
gradient budget, steps, or sample presentations, because half its OXE stream is replaced by deployment-distribution data at identical
total volume. A larger update or sample-presentation budget is
unavailable as an explanation; the recovery must arise from the
content of the substituted stream. Second, the repair stream is
small and reused: 38{,}880 frames revisited $\approx$13 times by
step 8{,}000. We disclose this; the held-out result below argues
against memorization of the recorded episodes. One argument is not
one scientific factor: the substituted stream bundles several
co-varying properties (Sec.~\ref{sec:threats}).

\subsection{Outcome: parity, reported against selection}
Checkpoint selection used the selection set only, where
step 8{,}000 was the in-band maximum (24/36), and was locked before
any held-out evaluation. Held-out was then evaluated once per
checkpoint for the three bracketing checkpoints: 20/36 (step 6k),
\textbf{18/36} (step 8k), 21/36 (step 10k). We report the
pre-selected checkpoint's 18/36 against the teacher's 17/36 under the identical harness, which we read as \emph{parity, not improvement} because the difference is within run-to-run repeatability (Sec.~\ref{sec:case}), and we decline the higher neighboring
number that post-hoc selection would have claimed. The selection-set
margin over the teacher (24/36 vs.\ 19/36, Table~\ref{tab:parity})
is consistent with specialization to the recorded scenes, because the selection IDs deterministically reproduce the very layouts on which the rollouts were recorded, which is one more reason the held-out
comparison, on never-recorded layouts, carries the report. Grasping does not
regress: held-out grasp 34/36, against the unrepaired student's 17/36, so the cure is not purchased by forgetting. The zero-control
run stays at floor at every checkpoint (single exception: 1/36 at
step 8k): the recovery is attributable to the substituted rollout
data rather than to the additional optimization, and which property
of that data does the work is the open question of
Sec.~\ref{sec:threats}.

\subsection{Recovery on every measured axis}
Parity of a single scalar can hide an unhealed policy. Two further
measurements say the recovery is complete in kind, not only in
rate, on every axis we measured.

\emph{The negative, late-heavy signature is eliminated.} On the
same paired 4-episode base as Sec.~\ref{sec:lesion}, the repaired
student's $d_z$ is $+0.0009$ all-phase and $-0.00065$ late, an order-of-magnitude reduction from V1's $-0.0067$ / $-0.0103$, and the early-third residual falls comparably
($-0.0069 \to -0.0005$). The mid-third, where the failing
signature was never consistently expressed, does not follow the
failing pattern (Table~\ref{tab:signature}). The repaired model
grasps 4/4 on a base where V1 managed 2/4.

\emph{The perturbation-response profile follows the teacher's.}
Under the same $z$-offset instrument of Sec.~\ref{sec:necessity},
the repaired student's counted dose-response curve shows a
dose-dependent degradation similar to the teacher's, with $22, 22, 14, 3, 0$ against $19, 16, 8, 3, 0$ over offsets $\{0, .002, .004, .006, .008\}$, while the zero-control stays at
floor throughout ($0, 2, 2, 1, 0$). The repaired model's higher
absolute counts at low doses mirror its zero-offset advantage on
these selection configurations (22 vs.\ 19), consistent with the
selection-set gap in Table~\ref{tab:parity}; the claim is the
shared degradation shape, not superiority. Parity extends from a
scalar to a behavioral profile: the healed policy responds to
perturbation like its teacher, not like its unhealed twin
(Table~\ref{tab:dose}).

\begin{table}[t]
\caption{\textbf{Dose response under command-level perturbation.}
Counted success (out of 36) under fixed additive $z$ offsets
applied throughout the rollout, evaluated on the
checkpoint-selection configurations (held-out configurations were
never used for perturbation sweeps). The teacher and the repaired
student show similar dose-dependent degradation; the zero-rollout
control and V1 remain at floor at every dose. The same intervention
that demonstrably degrades healthy policies does not rescue either
failing one (Secs.~\ref{sec:necessity},~\ref{sec:sufficiency}).
Zero-offset entries are separate perturbation-sweep evaluations,
not part of the checkpoint-selection comparison of
Table~\ref{tab:parity}.}
\label{tab:dose}
\centering
\footnotesize
\setlength{\tabcolsep}{2.5pt}
\begin{tabular}{lrrrrrrr}
\toprule
& \multicolumn{7}{c}{Fixed additive $z$ offset} \\
\cmidrule(lr){2-8}
Policy & 0 & .002 & .004 & .006 & .008 & .012 & .020 \\
\midrule
Teacher (12L)     & 19 & 16 &  8 & 3 & 0 & 0 & 0 \\
Repaired (8k)     & 22 & 22 & 14 & 3 & 0 & 0 & 0 \\
\midrule
Zero-control (8k) &  0 &  2 &  2 & 1 & 0 & 0 & 0 \\
Student V1        &  1 &  0 &  1 & 0 & 0 & 0 & 0 \\
\bottomrule
\end{tabular}
\end{table}

\begin{table}[t]
\caption{\textbf{Parity under a pre-committed selection rule.}
Step 8{,}000 was selected on the selection set only (36
configurations), where it was the in-band maximum, and locked
before any held-out evaluation; for the checkpoint-selection
comparison, the held-out set (36) was then evaluated once per
checkpoint. The pre-selected checkpoint scores 18/36 held-out
versus the teacher's 17/36 under an identical harness; the higher
neighboring 21/36 at step 10k is not used for reporting. Held-out
grasp (34/36 vs.\ the unrepaired student's 17/36) shows the
recovery does not trade away grasp performance.
$^\dagger$Pre-selected.}
\label{tab:parity}
\centering
\footnotesize
\begin{tabular}{lccc}
\toprule
& Selection & Held-out & Held-out \\
Policy & success & success & grasp \\
\midrule
Student V1 (8L, 300k)                & 1/36  & 0/36  & 17/36 \\
Teacher (12L)                        & 19/36 & 17/36 & 30/36 \\
\midrule
Repaired, step 6k                    & 20/36 & 20/36 & 34/36 \\
\textbf{Repaired, step 8k}$^\dagger$ & \textbf{24/36} & 18/36 & 34/36 \\
Repaired, step 10k                   & 20/36 & 21/36 & 33/36 \\
\bottomrule
\end{tabular}
\end{table}

\FloatBarrier
\subsection{The causal statement}
The interventions support a proximate cause stated at the system
level, while leaving its training-time formation and
representational origin unresolved:

\begin{quote}
The proximate cause of the student's closed-loop failure is a
deficiency of its learned, state-conditional action mapping,
expressed most strongly in the late transport regime and compounded
by closed-loop execution into categorical failure at the transport
stage.
\end{quote}

Each element is anchored as follows. \emph{State-conditional,
localized}: the graded-to-categorical stage profile (a policy that moves the object at 90\% and transports at 0\% has not lost capability uniformly) and the economy of the cure itself (38{,}880
frames at unchanged budget). \emph{Conditional, not marginal}:
Step~1 shifts the measured marginals without restoring success.
\emph{Not a single channel}: the most visible projection is the
small, negative, late-heavy $z$ residual of Sec.~\ref{sec:lesion},
compounded over the remaining rollout \cite{rossbagnell, dagger,
simchowitz}, yet the clamp shows the deficiency is not confined to
it. Nor is it explained by a fixed additive $z$ bias, additional
same-data training at tested budgets, an audited harness defect, or
an 8-layer capacity ceiling. Two controls bound compression's role:
the teacher, trained on the same offline corpus, never develops the
deficiency, and the same 8-layer network recovers to parity once
re-supervised. The compression-and-distillation procedure is
therefore an enabling condition, not a capacity verdict; the
representational route by which it enables is the layer we do not
reach. The state-conditional character and the closed-loop
amplification are system-level inferences from these interventions,
not direct measurements of the internal mapping.

The minimal pair supplies the positive causal evidence at this
functional level: under the matched protocol, substituting the
successful-teacher-rollout bundle restores held-out success to
teacher parity (18/36 vs.\ 17/36), eliminates the negative,
late-heavy trace signature, and recovers a teacher-like
perturbation-response profile, while the zero-rollout control
remains at floor. The failure is a repairable defect in the
learned policy, not an immutable architectural limitation. What
the design does not identify is the active component within the
bundle: the partial constraints of Sec.~\ref{sec:threats} disfavor
purely visual-domain, task-content, and reduced-OXE readings,
leaving deployment-state coverage, success-only filtering,
teacher-consistent labels, or their combination unresolved. We
read the pattern as most consistent with missing supervision on
the states the deployed student actually reaches, the classical covariate-shift account \cite{rossbagnell, dagger}, and label
this an interpretation, not a measurement.

The logic survives at this grain: the remedy names none of the
cause, and the cause, functionally stated, predicts the remedy
class. A controlled supplement establishes that a deficiency
exists and where it is expressed; it does not, by itself, name the
missing nutrient.

\section{Threats to Validity and Limitations}
\label{sec:threats}

\textbf{Simulator-artifact hypothesis.} Could the $z$-drift be a
contact-physics artifact rather than a policy pathology? Two
observations constrain this. The teacher runs under identical
contact physics and does not exhibit the paired signature; and the
cure operates entirely within the same simulator by changing training data alone, whereas a policy-independent physics artifact would
not be expected to disappear under data-only retraining. A residual
caveat remains: the student's earlier actions could steer it into
contact configurations where artifacts trigger preferentially. 
We therefore scope the claim to the policy under this simulator's
dynamics; hardware transfer is future work, for which a physical
WidowX platform is in place.

\textbf{Capacity.} Post-repair parity of the same 8-layer network
falsifies a capacity ceiling for this task. Capacity may still
shape which pathologies compression induces; that question is
representational.

\textbf{Representational origin.} What the pruned layers lost such
that the mapping fails on exactly these states is the layer we do
not reach; the anatomy is deliberately behavioral and systemic.

\looseness=-1 \textbf{The bundle confound.} The repair stream bundles four properties, namely deployment-distribution states, success-only filtering, teacher-consistent labels, and the simulator's visual domain, all of which fully co-vary in our data; $\rho_R$ additionally
covaries with reduced OXE share. The minimal pair establishes the
bundle as the difference-maker, not any component. Partial
constraints: stage-specificity with relatively preserved early
stages disfavors a purely visual-domain reading, and Step 1's
failure on genuine task-domain data disfavors a task-content
reading; the pure-OXE condition was sampled at two budgets (5k,
10k), both at floor, disfavoring the reduced-OXE reading.
Component-level arms (failure-rollout admixture, student-state
relabeling) are future work. The same success-only confound is
flagged untested in \cite{vlaad}; we disclose it.

\textbf{Measurement-scale limits.} Micro-displacement intervention
analysis at the 0.002--0.006 scale is not decision-grade at
attainable $n$ in this setting (Sec.~\ref{sec:case}); counted
outcomes and paired statistics carry all claims here. We commend
the distinction to evaluation practice.

\textbf{Scope.} One model family, one task, one simulator; claims
are existential accordingly. A cheap second testbed was sought and
not found: within this harness the teacher itself performs at 1/12
and 0/12 on the two candidate Bridge tasks (and 0/8 on a third),
so no in-family task could host the dissociation, itself an infrastructure observation: teacher competence bounds the supply of
cheap testbeds.

\section{Conclusion}

There exists a class of compression-induced closed-loop failure
that is invisible to a model family's own offline battery, graded
in its early stages and categorical at its final one, resistant to
symptomatic correction at every dose and to data-volume remedies at
matched budgets, and curable to teacher parity, with the
negative, late-heavy signature eliminated and a teacher-like
perturbation-response profile recovered, by substituting successful
teacher rollouts collected in the deployment environment into an
otherwise unchanged training stream. We established this at
single-case resolution with a four-rung necessity ladder and a
minimal-pair sufficiency arm verified from execution records. For
scalable robot-learning infrastructure the implication is
operational: offline gates are insufficient as standalone
acceptance tests for compressed policies, and the decisive evidence was cheap, since a few dozen closed-loop trials exposed the categorical
collapse, one paired trace comparison isolated its signature, and
together they found what every offline gate we applied missed. The
anatomy, protocols, and controls are offered as a template for the
next case.

\section*{Acknowledgment}
The author thanks Prof. Haijun Su (The Ohio State University) for helpful discussions, and the organizers and reviewers of the IROS 2026 ScaleInfra Workshop. This research received no external funding; compute and hardware were paid for with the author's personal funds.

\end{document}